\documentclass[runningheads]{llncs}
\usepackage[T1]{fontenc}
\usepackage{graphicx}
\usepackage{booktabs}
\usepackage[misc]{ifsym}

\usepackage{multirow}
\usepackage{multicol}
\usepackage[colorlinks=true, allcolors=blue]{hyperref}
\usepackage[table]{xcolor}      
\usepackage{arydshln}           
\usepackage{amssymb}
\usepackage{amsmath}
\usepackage{tikz}
\usetikzlibrary{shapes.geometric, arrows.meta, positioning, calc, backgrounds}
\definecolor{lightgray}{gray}{0.93}
\usepackage{wrapfig}
\usepackage{adjustbox}
\usepackage{float}
\usepackage{orcidlink}
\usepackage{tabularx}
\begin{document}

\title{Efficient Online Continual Foundation Model Fine-Tuning for Predictive Process Monitoring}

\titlerunning{Efficient Online Continual FM Fine-Tuning for PPM}

\author{Sjoerd van Straten \orcidlink{0009-0002-5772-9073}\inst{1} \and
Marwan Hassani \orcidlink{0000-0002-4027-4351}\inst{1,2}}
\authorrunning{S. van Straten and M. Hassani}
\institute{
  Department of Mathematics and Computer Science, Eindhoven University of Technology, The Netherlands\\
  \email{\{h.a.j.v.straten, m.hassani\}@tue.nl}
  \and
  Department of Mathematics and Computer Science, University of Wuppertal, Germany\\   \email{hassani@uni-wuppertal.de}}

\maketitle

\begin{abstract}
Predictive Process Monitoring (PPM) models are increasingly deployed in dynamic environments where concept drift causes the underlying process distribution to shift over time. While recent work has moved toward online continual learning, existing methods train compact, task-specific networks entirely from scratch, leaving a persistent cold-start problem. Foundation Models (FMs) offer a compelling solution to this problem, but their continual fine-tuning in the process mining domain remains unexplored. We propose COMPASS (Continual Online foundation Model-based PPM with Adaptive SubSpaces), the first framework for online continual fine-tuning of FMs for PPM. COMPASS adapts loss-plateau detection to autonomously identify task boundaries in event streams and maintains a unified knowledge subspace including both pre-trained and task-specific directions. We evaluate our approach on nine event streams covering synthetic and real-world concept drift scenarios, across task-free and task-aware settings with multiple backbones and with consistent hyperparameter tuning across all methods. Our approach outperforms three SOTA non-FM competitors and two update strategy baselines, with particularly strong gains on streams exhibiting recurrent drift and complex, long-running cases, while incurring acceptable computational overhead compared to the non-FM competitors.

\keywords{Foundation Models  \and Online Continual Learning \and Process Mining \and Predictive Process Monitoring \and Parameter-Efficient Fine-Tuning.}
\end{abstract}

\section{Introduction}\label{sec:introduction}

\begin{figure}[t]
    \centering
    \includegraphics[width=1\linewidth]{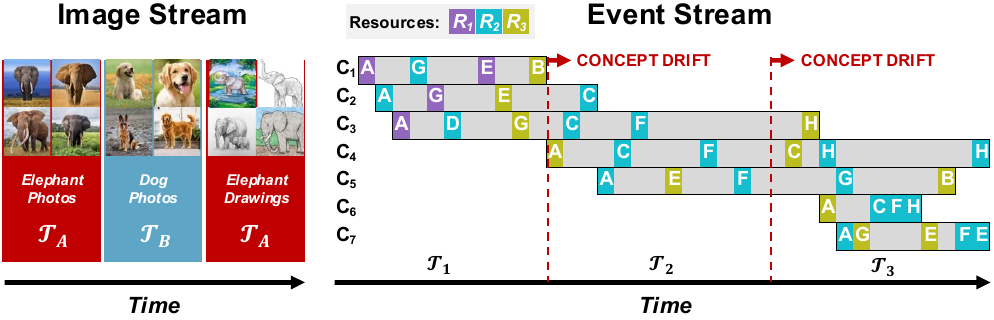}
    \caption{Comparison of an image stream and a process event stream under concept drift. In an image stream (\textit{left}) tasks are visually distinct between $\mathcal{T}_A$ and $\mathcal{T}_B$. In a process event stream (\textit{right}), cases ($C_1 - C_7$) are long-running and interleaved across task boundaries, and drift is caused by organizational changes causing activities (represented as letters) and the time between them to change. Resource R1 is on maternity leave after $\mathcal{T}_1$, and a policy change after $\mathcal{T}_2$ causes newly initiated cases to be prioritized, pushing older interleaved cases to the background. No input-space signal announces the shift, making autonomous adaptation to drift substantially harder than in image streams.}
    \label{fig:introduction}
\end{figure}
Predictive Process Monitoring (PPM) is a cornerstone of modern operational intelligence, enabling organizations to predict future process behavior, such as the next interaction in a customer journey with a company, for instance to proactively mitigate risks or optimize resource allocation~\cite{di2022predictive,ceravolo2024predictive}. Traditionally, PPM models were trained on static datasets under the assumption of a stable environment. However, real-world organizational processes are inherently dynamic and subject to concept drifts, where changes in regulations, seasonality or resource availability cause the underlying process distribution to vary over time. 

While recent advancements have moved toward online settings where models adapt to streaming event data~\cite{maisenbacher2017handling,rizzi2022update,marquez2022updating,EDBN}, current approaches suffer from at least one of two fundamental limitations. First, most existing methods rely on training compact, task-specific networks entirely from scratch for each new deployment~\cite{DynaTrainCDD,CNAPWP,TFCLPM}. This results in a persistent \textit{cold-start problem}, where the model possesses no prior sequence understanding and must learn the ``organizational logic'' of a process from scratch. Second, these models often struggle with the stability-plasticity dilemma, resulting from the need to adapt to new behaviors (plasticity) without overwriting or forgetting previously learned knowledge (stability), a phenomenon known as \textit{catastrophic forgetting}. 

These challenges are compounded by properties of process event streams that make online continual learning (OCL) substantially more complex than in domains such as image classification. As illustrated in Fig.~\ref{fig:introduction}, while an image stream presents clean observations with task boundaries that manifest as clear breaks in the input space, event streams are fundamentally different. An example of a \textit{case} is the connected treatment steps of a patient in a hospital or the end-to-end processing steps of a customer loan application in a bank. Each step occurrence is referred to as an \textit{event}, consisting of an \textit{activity} (e.g., the name of the treatment step) that is executed by a \textit{resource} (e.g., the doctor) at a certain \textit{timestamp}. Cases are long-running and interleaved across task boundaries, while drift is caused by organizational changes such as policy updates, staffing change or seasonality that are not necessarily reflected in the events themselves. The only evidence of a shift is a gradual change in the distribution of activities, resources and time between activities, in addition to decision points. Together with the interleaved cases, this makes autonomous drift adaptation a noisier and more delayed problem than in other domains. 

Foundation Models (FMs), especially Large Language Models (LLMs), offer a compelling response to the cold-start problem. Pre-trained on a large corpora of text, they encode sequence-modeling capabilities that transfer naturally to process traces when activities are treated as discrete tokens~\cite{oyamada2025domain}. PPM tasks include for example next activity prediction (e.g., predicting the next treatment step for a patient), final outcome prediction (e.g., whether their treatment will be satisfying) and remaining time prediction (e.g., the time until the patient leaves the hospital). In this work we focus on next activity prediction as a representative task. We expect the proposed framework to extend to other PPM tasks, though we do not evaluate this empirically. Unlike NLP and computer vision, where domain-specific FMs are well-established, no FMs exist for PPM. Therefore, FMs must be adapted to the process mining domain on the fly. Parameter-efficient fine-tuning (PEFT) via LoRA~\cite{hu2022lora} makes FM adaptation efficient and computationally feasible, but vanilla LoRA provides no mechanism for preserving either pre-trained or previously learned task knowledge as the stream evolves.

We propose COMPASS (\textbf{C}ontinual \textbf{O}nline foundation 
\textbf{M}odel-based \textbf{P}PM with \textbf{A}daptive 
\textbf{S}ub\textbf{S}paces), the first framework for online continual fine-tuning of FMs for next activity prediction. Upon detecting that the loss has stabilized, signaling that the current task distribution has been captured, our approach expands a unified knowledge space including both the pre-trained FM's dominant input directions and the task-specific directions accumulated across all observed tasks. New adapters are reinitialized into the residual subspace orthogonal to the unified knowledge space, ensuring plasticity toward new tasks without interference with prior knowledge.

The contributions of this work are threefold. (i) we adapt the loss-plateau detection of~\cite{wei2025online} to process event streams for autonomous task adaptation; (ii) we adapt and combine the residual subspace projection of~\cite{luo2026keeplora} with pre-trained knowledge anchoring to jointly achieve plasticity, backward stability and forward stability in event streams; and (iii) we evaluate our approach on nine event streams under synthetic and real-world concept drifts, across task-free and task-aware settings with multiple backbones, demonstrating state-of-the-art performance on next activity prediction against three non-FM competitors and two update strategy baselines. Our contribution is primarily the integration and adaptation of these mechanisms into a coherent framework for the PPM domain, rather than the introduction of new continual learning algorithms in isolation.

The remainder of this paper is structured as follows. Section~\ref{sec:preliminaries_problem_definition} introduces the preliminaries and formally defines the problem. Section~\ref{sec:related_work} reviews related work and defines the research gap. Section~\ref{sec:method} describes the proposed framework in detail. Section~\ref{sec:experimental_setup} and Section~\ref{sec:evaluation_results} present the experimental setup and results, respectively. Section~\ref{sec:conclusion} concludes with a discussion and avenues for future work.

\section{Preliminaries and Problem Definition}\label{sec:preliminaries_problem_definition}

Let $\mathcal{A}$ denote the universe of all possible process activities. The execution of a process instance yields a \textit{trace}, defined as a finite sequence $\sigma = \langle a_1, a_2, \ldots, a_L \rangle \in \mathcal{A}^*$ of length $L$, where each element $a_i \in \mathcal{A}$ represents the activity label of the $i$-th event. In this work, we focus on next activity prediction, which is defined over prefixes of such traces as follows:

\begin{definition}[Next Activity Prediction]
Given a prefix $x = \langle a_1, a_2, \ldots, a_k \rangle$ of length $1 \leq k < L$,
the next activity prediction task is to predict $\hat{a}_{k+1}$ that should
ideally match the ground truth target $y = a_{k+1}$.
\end{definition}

In an online setting, events arrive as a continuous, non-stationary event stream $S$ rather than a static dataset. All prefix-target pairs $(x_i, y_i)$ are sorted by timestamp and partitioned into contiguous, non-overlapping windows $\langle \gamma_1, \gamma_2, \ldots \rangle$ where each $\gamma_l = \{(x_i, y_i)\}_{i=1}^{\Gamma}$ is a batch of $\Gamma$ pairs. Due to concept drifts, the underlying joint distribution $P(X,Y)$ may shift over time. A \textit{task} is defined as a maximal contiguous subsequence of $\mathcal{S}$ over which $P(X, Y)$ remain stable. The model $f_\theta$ must predict before updating via a \textit{test-then-train} protocol, preventing data leakage. We consider a model $f_\theta$ parameterized by $\theta$, where the base weights $W_0 \in \mathbb{R}^{d \times m}$ with hidden dimension $d$ and output dimension $m$ encode general sequence understanding acquired during large-scale pre-training on text corpora. Rather than updating all of $\theta$, parameter-efficient fine-tuning constrains adaptation to a low-dimensional subspace via an increment $\Delta W$ with $\|\Delta W\|_0 \ll dm$, leaving $W_0$ nominally frozen. The model's objective over window $\gamma_l$ is to minimize the cross-entropy loss (Eq.~\ref{eq:ce_loss}):

\begin{equation} \label{eq:ce_loss}
    \mathcal{L}_{\text{CE}}(\gamma_l) = -\frac{1}{\Gamma} \sum_{(x_i, y_i) \in \gamma_l} \log P(y_i \mid x_i;\, W_0 + \Delta W)
\end{equation}

In continual learning, the model observes tasks $\mathcal{T}_1, \mathcal{T}_2, \ldots$ sequentially and must maintain performance across all tasks seen so far. The primary failure mode is \textit{catastrophic forgetting}. Updating $\theta$ to minimize loss on $\mathcal{T}_t$ may degrade performance on $\mathcal{T}_{t'}$ for $t' < t$. In addition, aggressive updates risk overwriting the general sequence understanding in $W_0$, degrading the model's ability to generalize to unseen process behavior. In the \textit{task-free} setting considered in this work, task boundaries are unknown a priori and must be inferred from the event stream itself.

This gives rise to the following problem. Given a pre-trained FM $f_\theta$ with base weights $W_0 \in \mathbb{R}^{d \times m}$ 
and a non-stationary event stream $\mathcal{S}$ with unknown task boundaries, find a 
parameter update strategy that, for all tasks $\mathcal{T}_1, \mathcal{T}_2, \ldots$ 
observed so far, jointly satisfies:
\begin{itemize}
    \item \textbf{Plasticity}: $f_\theta$ minimizes $\mathcal{L}_{\text{CE}}$ on the 
    current task $\mathcal{T}_t$;
    \item \textbf{Backward stability}: performance degradation on all previously observed tasks 
    $\mathcal{T}_{t'}$, $t' < t$, is minimized;
    \item \textbf{Forward stability}: the general sequence understanding encoded in 
    $W_0$ is preserved throughout the stream.
\end{itemize}

\section{Related Work}\label{sec:related_work}

\subsubsection{Update Strategies.} \label{sec:updating_strategies}
Early work on model maintenance in PPM focused on periodic offline retraining. Maisenbacher and Weidlich~\cite{maisenbacher2017handling} formalize concept drift in PPM and present a paradigm grounded in incremental learning, showing that advanced drift-aware classifiers are best suited for continuously evolving processes. Márquez-Chamorro et al.~\cite{marquez2022updating} compare six data selection strategies for offline model rebuilding, finding that ensemble and non-cumulative approaches outperform naive full retraining, with gains of up to fifteen percentage points over a no-update baseline. Rizzi et al.~\cite{rizzi2022update} similarly evaluate four update strategies for final outcome prediction, demonstrating that incremental updates match full retraining accuracy at a fraction of the computational cost. Pauwels and Calders~\cite{EDBN} extend this to next activity prediction, showing that incremental strategies can exploit catastrophic forgetting as a natural mechanism for discarding outdated process behavior. These works establish that continual adaptation is essential in dynamic environments. However, they all assume periodic, batch-oriented updates rather than true event-by-event online streaming and train task-specific models entirely from scratch, foregoing the structural knowledge and sequence understanding that FMs can provide out of the box.

\subsubsection{Online Continual Learning Approaches for PPM.} \label{sec:ocl_for_ppm}
A second generation of methods moves toward genuinely online settings where models adapt to streaming event data without periodic rebuilding. DynaTrainCDD~\cite{DynaTrainCDD} monitors the incoming stream using the PrefixCDD~\cite{huete2023prefixcdd} algorithm, triggering retraining upon detected structural deviations, though it reinitializes a single dense layer (SDL) backbone~\cite{EDBN} from scratch at each drift point and does not address catastrophic forgetting. TFCLPM~\cite{TFCLPM} takes a task-free continual learning perspective, combining a dynamic loss function with Memory Aware Synapses and a diversity-aware hard sample buffer to mitigate forgetting, while still relying on a lightweight SDL network trained from scratch. CNAPwP~\cite{CNAPWP} adapts the DualPrompt~\cite{wang2022dualprompt} architecture to next activity prediction, using learnable general and expert prompts to disentangle task-invariant from task-specific knowledge, demonstrating strong performance under recurrent concept drifts. However, these existing approaches share a fundamental limitation. They train compact, task-specific networks entirely from scratch for each new deployment. This means the cold-start problem persists, each new event stream must be learned from the ground up, without any prior sequence understanding. FMs, by contrast, arrive with generalizable sequence knowledge that can be efficiently adapted rather than rebuilt, opening a path toward efficient online continual learning for PPM.  

\subsubsection{Continual Fine-Tuning of Foundation Models.}
Low-Rank Adaptation (LoRA)~\cite{hu2022lora} provides the foundation for parameter-efficient fine-tuning by freezing base weights and injecting pairs of trainable low-rank matrices into transformer layers, reducing trainable parameters by orders of magnitude while introducing no additional inference latency. Building on this, several works have explored LoRA fine-tuning in continual learning settings. Online-LoRA~\cite{wei2025online} extends LoRA to the task-free online setting for vision transformers, using loss-surface plateaus to autonomously detect distribution shifts and trigger initialization of new LoRA parameters, combined with an online weight regularization scheme to mitigate catastrophic forgetting. KeepLoRA~\cite{luo2026keeplora} further addresses the preservation of general pre-trained knowledge by constraining LoRA updates to the residual subspace of the parameter space, thereby jointly achieving forward stability, backward stability and plasticity. Despite these advances in computer vision and NLP domains, the continual fine-tuning of Foundation Models in the context of process mining remains unexplored, representing a clear research gap that this work addresses.
\section{Method} \label{sec:method}

\begin{figure}[h]
    \centering
    \includegraphics[width=1\linewidth]{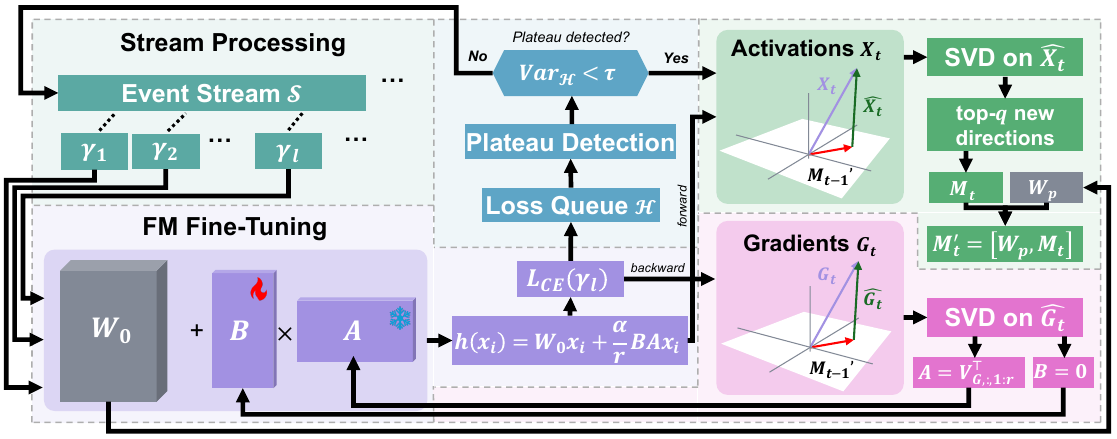}
    \caption{Overview of the COMPASS architecture. Colors indicate functional stages: LoRA fine-tuning (purple), plateau detection (blue), knowledge subspace expansion (green), and adapter reinitialization (pink), described in order below.}
    \label{fig:architecture}
\end{figure}

We propose COMPASS (\textbf{C}ontinual \textbf{O}nline foundation \textbf{M}odel-based \textbf{P}PM with \textbf{A}daptive \textbf{S}ub\textbf{S}paces), a framework for online continual next activity prediction using FMs (cf. Fig.~\ref{fig:architecture}, where colors indicate the functional stages which are described in detail below). Process activities are tokenized directly as integers via $\phi : \mathcal{A} \rightarrow \mathbb{Z}$ following~\cite{oyamada2025domain}, and each prefix $x_i = \langle a_1, \ldots, a_k \rangle$ is padded to length $L_{\max}$. Prefix-target pairs are sorted by timestamp into $S$ and partitioned into non-overlapping windows $\gamma_l$ of size $\Gamma$. For each $\gamma_l$, the model first predicts all events and then updates its weights, preventing data leakage. The FM $f_{\theta}$ is adapted via LoRA~\cite{hu2022lora} (purple stage in Fig.~\ref{fig:architecture}). Base weights $W_0 \in \mathbb{R}^{d \times m}$ are frozen and low-rank matrices $A \in \mathbb{R}^{r \times m}$ and  $B \in \mathbb{R}^{d \times r}$ are injected, giving the modified forward pass (Eq.~\ref{eq:forward}):

\begin{equation} \label{eq:forward}
    h(x_i) = W_0 x_i + \frac{\alpha}{r} B A x_i
\end{equation}

where $\alpha = 2r$ is a scaling factor. The model is optimized with cross-entropy loss $\mathcal{L}_{\text{CE}}(\gamma_l)$ over each window. This reduces trainable parameters by orders of magnitude and makes real-time adaptation on event streams computationally feasible. 

\subsubsection{Plateau Detection.}\label{sec:drift_detection} Since task boundaries are unknown a priori, we detect them via a rolling variance $\text{Var}_{\mathcal{H}}$ of a queue $\mathcal{H}$ of the $N$ most recent window losses (blue stage in Fig.~\ref{fig:architecture}), adapted from~\cite{wei2025online}. When $\text{Var}_{\mathcal{H}} < \tau$, the loss has stabilized which indicates that the task distribution has been captured by $f_{\theta}$. This triggers the knowledge consolidation phase. A cooldown period of $C$ windows prevents redundant triggers within the same task.

\subsubsection{Knowledge Subspace Expansion.} Upon detecting a plateau for task $\mathcal{T}_t$, COMPASS expands a unified knowledge subspace (green stage in Fig.~\ref{fig:architecture}) to incorporate the directions most relevant to task $\mathcal{T}_t$ adapted from~\cite{luo2026keeplora}. This subspace has two components consisting of a fixed anchor derived from $W_0$ and a growing set of task-specific directions accumulated across all observed tasks.\\

For clarity, Table~\ref{tab:notation} summarizes the subspaces introduced below: $W_p$ (fixed, pre-trained directions), $M_t$
(task-specific directions accumulated up to task $t$), and $M'_t = [W_p, M_t]$ (the unified subspace used for orthogonal projection).

\begin{table}[t]
\centering
\caption{Summary of the subspaces used in knowledge subspace expansion.}
\label{tab:notation}
\begin{tabularx}{\textwidth}{@{}l X l@{}}
\toprule
\textbf{Symbol} & \textbf{Meaning} & \textbf{Updated} \\
\midrule
$W_p$ & Dominant input directions of pre-trained $W_0$ & Fixed after initial SVD \\
$M_t$ & Task-specific directions accumulated across tasks $1,\dots,t$ & Grows at each consolidation \\
$M'_t = [W_p, M_t]$ & Unified knowledge subspace used for orthogonal projection & Grows, capped at rank $R$ \\
\bottomrule
\end{tabularx}
\end{table}

\paragraph{Pre-Trained Knowledge Subspace $W_p$:} The pre-trained FM $f_\theta$ encodes general sequence understanding acquired during large-scale language pre-training. Overwriting these capabilities during continual fine-tuning causes forward forgetting, degrading the model's ability to generalize to unseen processes. To preserve this knowledge, for each LoRA-wrapped layer we decompose $W_0 = USV^\top$ and retain the top-$p$ right singular vectors (spanning the dominant input directions of $W_0$) as columns of $W_p \in \mathbb{R}^{m \times p}$, such that:

\begin{equation}
    \|W_p\|_F^2 \;\geq\; \epsilon_w\,\|W_0\|_F^2
\end{equation}

where $\|\cdot\|_F$ denotes the Frobenius norm and $\epsilon_w \in (0,1)$  is the energy retention threshold. All subsequent LoRA updates are constrained to be orthogonal to $W_p$, ensuring the pre-trained sequence understanding remains intact throughout the event stream. 

\paragraph{Previous Task Knowledge Subspace $M_t$:} Forward hooks capture the input activations $X_t$ at each LoRA-wrapped layer, collecting all token-level representations from window $\gamma_l$. Novel task-specific directions are isolated by projecting onto the
residual subspace orthogonal to both $W_p$ and the previously accumulated directions $M_{t-1} \in \mathbb{R}^{m \times n_{t-1}}$, where $n_{t-1}$ is the number of task-specific directions stored after $t-1$ consolidations:

\begin{equation}
    \hat{X}_t \;=\; X_t \;-\; \underbrace{X_t W_p W_p^\top}_{\text{pre-trained directions}} \;-\; \underbrace{X_t M_{t-1} M_{t-1}^\top}_{\text{previous task directions}}
\end{equation}

SVD on $\hat{X}_t$ yields the top-$q$ dominant singular right vectors, where $q$ is determined dynamically by energy threshold $\epsilon_f \in (0,1)$, retaining the minimum number of directions needed to account for fraction $\epsilon_f$ of the residual energy. These new directions are appended to $M_{t-1}$ and the combined matrix is re-orthogonalized via a secondary SVD, resulting in $M_t$. $W_p$ and $M_t$ are then unified into a single consolidated subspace $M'_t = [W_p, M_t] \in \mathbb{R}^{m \times (p+n_t)}$, which grows monotonically across task boundaries and serves as a compressed memory of all process knowledge encountered so far. In practice, the number of columns retained in both $W_p$ and $M_t$ is capped at a maximum rank $R$ for memory efficiency, bounding the growth of $M_t'$ across task boundaries.

\subsubsection{Adapter Reinitialization.} With $M_{t}'$ updated, the LoRA adapter is reinitialized (pink stage in Fig.~\ref{fig:architecture}) to: (i) span a subspace orthogonal to $M_{t-1}'$, preventing interference with prior knowledge, and, (ii) approximate the gradient of the new task, maintaining plasticity. We temporarily unfreeze $W_0$ and compute the full-weight gradient $G_t = \nabla_{W_0}\mathcal{L}(W_0;\,\gamma_l) \in \mathbb{R}^{d \times m}$ from a single forward-backward pass on the current window. This gradient is projected onto the residual subspace orthogonal to $M_{t-1}'$:

\begin{equation}
    \hat{G}_t \;=\; \underbrace{G_t}_{\text{plasticity}} \;-\; 
\underbrace{G_t W_p W_p^\top - G_t M_{t-1} M_{t-1}^\top}_{\text{stability}}
\end{equation}

SVD on $\hat{G}_t = U_G S_G V_G^\top$ yields the top-$r$ singular vectors, which initialize the LoRA matrices $A = V_{G,\,:,1:r}^\top$ and $B = \mathbf{0}$. $A$ captures the dominant directions of change required for the new task, while $B = \mathbf{0}$ delays their effect until training updates $B$. $A$ is then frozen and $B$ is trained for subsequent windows. Since $\frac{\alpha}{r}BA = \mathbf{0}$ at initialization, the forward pass is unchanged at the moment of consolidation, which ensures a stable transition between tasks. Optimizing $B$ alone is equivalent to gradient descent on $W_0$ constrained to span($A$)~\cite{luo2026keeplora} (a subspace orthogonal by construction to $M_{t-1}'$). We thereby
ensure that new learning prevents overwriting previously learned process knowledge, mitigating catastrophic forgetting effectively.

\section{Experimental Setup}\label{sec:experimental_setup}

\subsubsection{Datasets and Metrics.}
We evaluate our approach on nine event streams covering synthetic and real-world concept drift scenarios. The five synthetic streams (\textit{IRO5000}, \textit{ORI5000}, \textit{ROI5000}, \textit{OIR5000} and \textit{RIO5000})~\cite{maaradji2015fast} simulate recurrent drifts to test the capacity to retain past knowledge. For real-world evaluation we use four streams from the BPI Challenges\footnote{The BPI Challenge datasets are publicly available at \url{https://data.4tu.nl}}. We use three subsets of the BPIC 2020, named DomesticDeclarations (\textit{BPI20-DD}), InternationalDeclarations (\textit{BPI20-ID}) and RequestForPayment (\textit{BPI20-RFP)}. Additionally, we evaluate on a recurrent version of BPIC 2015 (\textit{BPI15-REC)}, constructed from the first three municipalities, each recurring three times. We report average accuracy over the full evaluation stream and per-window accuracy at given index, providing both a holistic and fine-grained view of performance under concept drift. For computational efficiency, we report end-to-end runtime and memory consumption over the stream.

\subsubsection{Baselines.} We compare against five baselines consisting of two update strategies and three SOTA non-FM competitors.\textbf{\textit{ DoNothing~\cite{rizzi2022update}}} trains only on the first three windows and never updates, serving as a lower bound. \textbf{\textit{$\gamma$ = LastDrift~\cite{EDBN}}} retrains monthly on all data since the most recent drift. \textbf{\textit{DynaTrainCDD~\cite{DynaTrainCDD}}} monitors the stream with PrefixCDD~\cite{huete2023prefixcdd} and triggers retraining on structural deviations. \textbf{\textit{TFCLPM~\cite{TFCLPM}}} is a task-free online continual learning framework using a single dense layer (SDL) network with dynamic loss function and a diversity-aware hard-sample buffer. \textbf{\textit{CNAPwP~\cite{CNAPWP}}} adopts the DualPrompt~\cite{wang2022dualprompt} architecture with learnable general and expert prompts for recurrent drift handling. 

\subsubsection{Implementation Details.} COMPASS is evaluated with both the Tiny-LLM~\footnote{\url{https://huggingface.co/arnir0/Tiny-LLM}} (10M parameters) and DistilGPT2~\footnote{\url{https://huggingface.co/distilbert/distilgpt2}} (82M parameters) backbones. To isolate the contribution of the continual learning strategy, DoNothing and $\gamma$ = LastDrift are evaluated with the DistilGPT2 backbone, which represents the best-performing COMPASS configuration. We further distinguish a task-aware variant (oracle drift boundaries) from our proposed task-free variant (plateau detection, cf. Section~\ref{sec:drift_detection}). Experiments have been run on an HPC cluster with NVIDIA Tesla V100 (16GB) GPUs. We perform 5 repetitions with different seeds, reporting mean and standard deviation. To assess whether performance differences are meaningful, we conduct a paired two-sided student's t-test ($\alpha = 0.05$) between our best-performing COMPASS configuration and the best competing baseline on each experiment, using the per-seed accuracies as paired samples. The first 15\% of cases of each dataset serve as a validation set for hyperparameter tuning, the remaining 85\% is reserved for online evaluation following the test-then-train protocol, preventing data leakage. The full implementation including key parameters and additional results for each method and dataset is available in the public GitHub repository~\footnote{\url{https://github.com/SvStraten/COMPASS}}.

\section{Evaluation Results}\label{sec:evaluation_results}

\subsubsection{Average Accuracy.} Table~\ref{tab:average_accuracy} reports average accuracy across all methods and datasets. DoNothing~\cite{rizzi2022update} fails to keep up with the stream, achieving near-zero accuracy on BPI15-REC and remaining far below all adaptive methods, confirming that a static model is infeasible under concept drift. Our approach with the DistilGPT2 backbone consistently achieves the highest next activity prediction accuracy, outperforming all baselines on seven of the nine evaluated logs. Gains are the highest on the recurrent BPI15-REC dataset, where COMPASS DistilGPT2 substantially outperforms the next best baseline CNAPwP~\cite{CNAPWP} by +19\%, highlighting the benefit of our approach on streams with complex, long-running cases. Strong performance on the synthetic streams, which simulate recurrent drift, further demonstrate that COMPASS retains previously learned task knowledge and recovers it when earlier process behaviors reappear, rather than overwriting it with each new adaptation. On the BPIC 2020 streams, differences are smaller as these logs exhibit less severe drift. The smaller Tiny-LLM backbone performs similar to the DistilGPT2 backbone, but at a fraction of the parameters. Crucially, the task-free variant closely tracks the task-aware oracle across all datasets, demonstrating that plateau  detection is a reliable substitute when known task boundaries are unknown. 

\begin{table}[t]
\caption{Average accuracy for all baselines and datasets. \textbf{Bold} denotes the highest accuracy, \textit{italic} and \underline{underlined} the second and third highest. Sig. = statistically significant difference ($\alpha$ = .05).}
\label{tab:average_accuracy}
\setlength{\tabcolsep}{2pt}
\renewcommand{\arraystretch}{1.3}
\resizebox{\textwidth}{!}{%
\begin{tabular}{lccccccccc}
\toprule
\textbf{Method}
& \rotatebox{60}{\textit{IRO5000}}
& \rotatebox{60}{\textit{ORI5000}}
& \rotatebox{60}{\textit{ROI5000}}
& \rotatebox{60}{\textit{OIR5000}}
& \rotatebox{60}{\textit{RIO5000}}
& \rotatebox{60}{\textit{BPI15-REC}}
& \rotatebox{60}{\textit{BPI20-RFP}}
& \rotatebox{60}{\textit{BPI20-DD}}
& \rotatebox{60}{\textit{BPI20-ID}} \\
\midrule
\textbf{DoNothing}~\cite{rizzi2022update}
  & .192\textcolor{gray}{\tiny$\pm$.107}
  & .200\textcolor{gray}{\tiny$\pm$.068}
  & .223\textcolor{gray}{\tiny$\pm$.097}
  & .205\textcolor{gray}{\tiny$\pm$.072}
  & .171\textcolor{gray}{\tiny$\pm$.072}
  & .013\textcolor{gray}{\tiny$\pm$.006}
  & .511\textcolor{gray}{\tiny$\pm$.088}
  & .495\textcolor{gray}{\tiny$\pm$.090}
  & .247\textcolor{gray}{\tiny$\pm$.008} \\
\midrule
\textbf{$\gamma$ = LastDrift}~\cite{EDBN}
  & .220\textcolor{gray}{\tiny$\pm$.085}
  & .770\textcolor{gray}{\tiny$\pm$.003}
  & .246\textcolor{gray}{\tiny$\pm$.056}
  & .709\textcolor{gray}{\tiny$\pm$.003}
  & .269\textcolor{gray}{\tiny$\pm$.052}
  & .484\textcolor{gray}{\tiny$\pm$.000}
  & \textbf{.886\textcolor{gray}{\tiny$\pm$.001}}
  & .863\textcolor{gray}{\tiny$\pm$.001}
  & .840\textcolor{gray}{\tiny$\pm$.006} \\
\textbf{DynaTrainCDD}~\cite{DynaTrainCDD}
  & .775\textcolor{gray}{\tiny$\pm$.005}
  & .785\textcolor{gray}{\tiny$\pm$.006}
  & .790\textcolor{gray}{\tiny$\pm$.001}
  & .729\textcolor{gray}{\tiny$\pm$.004}
  & .784\textcolor{gray}{\tiny$\pm$.004}
  & .455\textcolor{gray}{\tiny$\pm$.008}
  & .831\textcolor{gray}{\tiny$\pm$.010}
  & .830\textcolor{gray}{\tiny$\pm$.011}
  & .793\textcolor{gray}{\tiny$\pm$.003} \\
\textbf{TFCLPM}~\cite{TFCLPM}
  & \textit{.803\textcolor{gray}{\tiny$\pm$.001}}
  & \underline{.817\textcolor{gray}{\tiny$\pm$.002}}
  & \underline{.825\textcolor{gray}{\tiny$\pm$.011}}
  & .775\textcolor{gray}{\tiny$\pm$.002}
  & \underline{.814\textcolor{gray}{\tiny$\pm$.001}}
  & .473\textcolor{gray}{\tiny$\pm$.003}
  & .863\textcolor{gray}{\tiny$\pm$.009}
  & .876\textcolor{gray}{\tiny$\pm$.003}
  & .820\textcolor{gray}{\tiny$\pm$.002} \\
\textbf{CNAPwP}~\cite{CNAPWP}
  & \underline{.802\textcolor{gray}{\tiny$\pm$.002}}
  & .816\textcolor{gray}{\tiny$\pm$.005}
  & \textit{.829\textcolor{gray}{\tiny$\pm$.002}}
  & \underline{.780\textcolor{gray}{\tiny$\pm$.001}}
  & .812\textcolor{gray}{\tiny$\pm$.006}
  & \underline{.492\textcolor{gray}{\tiny$\pm$.015}}
  & \underline{.879\textcolor{gray}{\tiny$\pm$.001}}
  & \underline{.881\textcolor{gray}{\tiny$\pm$.008}}
  & \underline{.846\textcolor{gray}{\tiny$\pm$.003}} \\
\midrule
\multicolumn{10}{l}{\small\textit{\textcolor{gray}{Task-Aware}}} \\[-2pt]
\textbf{COMPASS}\textsubscript{Tiny-LLM}
  & .794\textcolor{gray}{\tiny$\pm$.007}
  & \textbf{.827\textcolor{gray}{\tiny$\pm$.002}}
  & .821\textcolor{gray}{\tiny$\pm$.011}
  & \textbf{.787\textcolor{gray}{\tiny$\pm$.001}}
  & .793\textcolor{gray}{\tiny$\pm$.016}
  & .582\textcolor{gray}{\tiny$\pm$.019}
  & \underline{.879\textcolor{gray}{\tiny$\pm$.002}}
  & \underline{.890\textcolor{gray}{\tiny$\pm$.001}}
  & .843\textcolor{gray}{\tiny$\pm$.002} \\
\textbf{COMPASS}\textsubscript{DistilGPT2}
  & \textbf{.812\textcolor{gray}{\tiny$\pm$.002}}
  & \underline{.825\textcolor{gray}{\tiny$\pm$.002}}
  & \textbf{.840\textcolor{gray}{\tiny$\pm$.002}}
  & \underline{.784\textcolor{gray}{\tiny$\pm$.001}}
  & \textbf{.821\textcolor{gray}{\tiny$\pm$.001}}
  & \textbf{.681\textcolor{gray}{\tiny$\pm$.002}}
  & \textbf{.886\textcolor{gray}{\tiny$\pm$.001}}
  & \textbf{.895\textcolor{gray}{\tiny$\pm$.001}}
  & \textbf{.851\textcolor{gray}{\tiny$\pm$.001}} \\
\multicolumn{10}{l}{\small\textit{\textcolor{gray}{Task-Free}}} \\[-2pt]
\textbf{COMPASS}\textsubscript{Tiny-LLM}
  & .787\textcolor{gray}{\tiny$\pm$.013}
  & \textit{.826\textcolor{gray}{\tiny$\pm$.001}}
  & .823\textcolor{gray}{\tiny$\pm$.024}
  & \textit{.786\textcolor{gray}{\tiny$\pm$.001}}
  & .803\textcolor{gray}{\tiny$\pm$.014}
  & \underline{.585\textcolor{gray}{\tiny$\pm$.009}}
  & \underline{.879\textcolor{gray}{\tiny$\pm$.002}}
  & \underline{.890\textcolor{gray}{\tiny$\pm$.002}}
  & .840\textcolor{gray}{\tiny$\pm$.001} \\
\textbf{COMPASS}\textsubscript{DistilGPT2}
  & \textbf{.812\textcolor{gray}{\tiny$\pm$.001}}
  & \textit{.826\textcolor{gray}{\tiny$\pm$.001}}
  & \textbf{.840\textcolor{gray}{\tiny$\pm$.001}}
  & .783\textcolor{gray}{\tiny$\pm$.002}
  & \textit{.820\textcolor{gray}{\tiny$\pm$.001}}
  & \textit{.676\textcolor{gray}{\tiny$\pm$.002}}
  & \textit{.885\textcolor{gray}{\tiny$\pm$.002}}
  & \textit{.893\textcolor{gray}{\tiny$\pm$.001}}
  & \textit{.849\textcolor{gray}{\tiny$\pm$.001}} \\
\midrule
\textbf{Sig.\ ($\alpha=.05$)} & * & * & * & * & * & * & - & * & * \\
\bottomrule
\end{tabular}%
}
\end{table}

\subsubsection{Accuracy at Given Index.} 
\begin{figure}[h]
    \centering
    \includegraphics[width=1\linewidth]{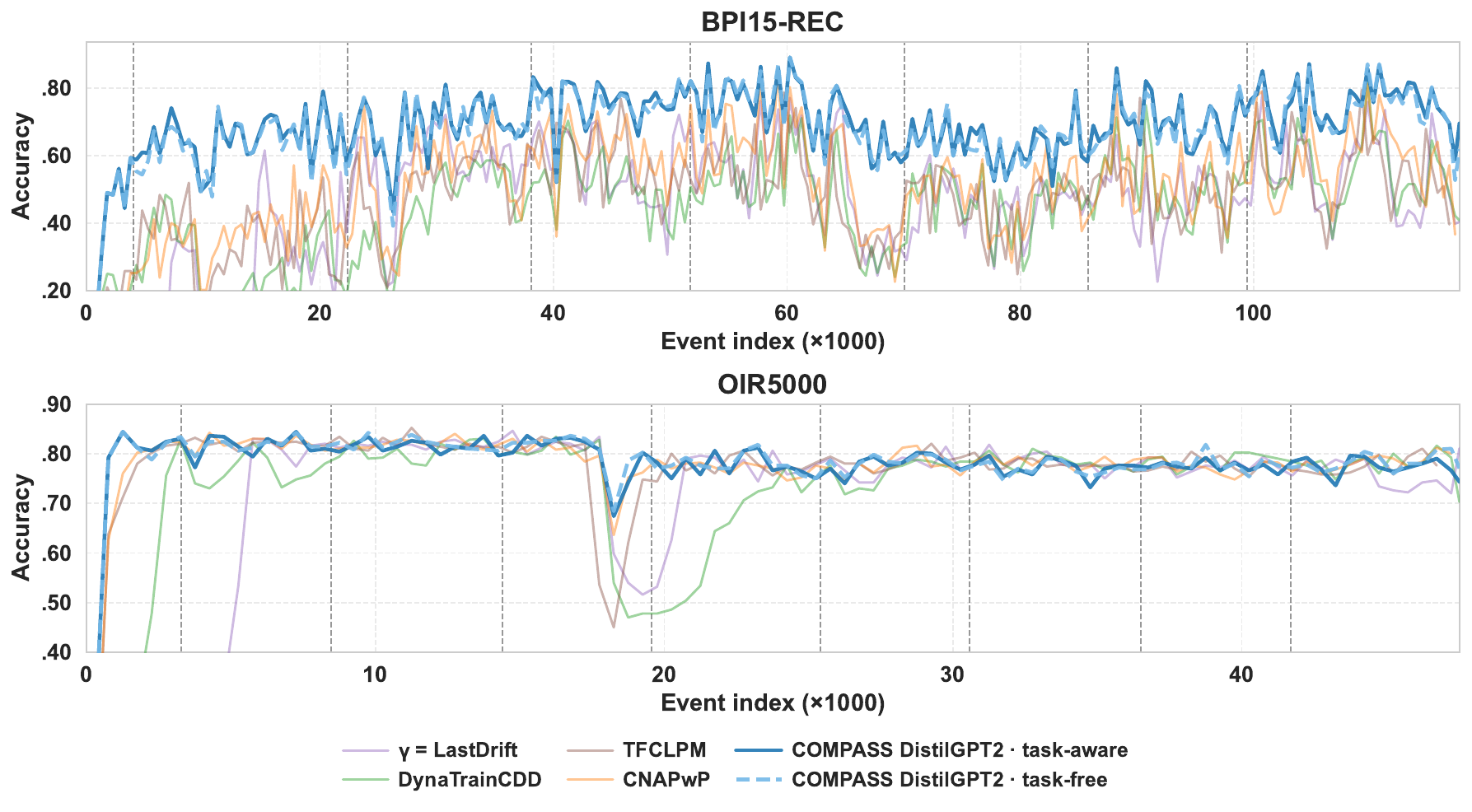}
    \caption{Accuracy at given index for the BPI15-REC and OIR5000 stream, illustrating COMPASS' faster recovery after concept drift and mitigation of the cold-start problem.}
    \label{fig:stream_accuracy}
\end{figure}

Fig.~\ref{fig:stream_accuracy} shows the stream accuracy over the BPI15-REC and ORI5000 stream, averaged per window. Dashed vertical lines show known concept drifts. For visual clarity, we show the DistilGPT2 variants only and remove the DoNothing~\cite{rizzi2022update}. From the very start of the evaluation stream, COMPASS achieves higher accuracy than the baselines, reflecting the advantage of pre-trained sequence understanding over randomly initialized models that must learn process behavior from scratch. COMPASS further maintains consistently higher accuracy throughout the streams, recovering more quickly after each concept drift. The task-free and task-aware variants follow nearly identical trajectories, further validating the plateau detection mechanism.

\subsubsection{Computational Efficiency.} 
\begin{figure}[t]
    \centering
    \includegraphics[width=1\linewidth]{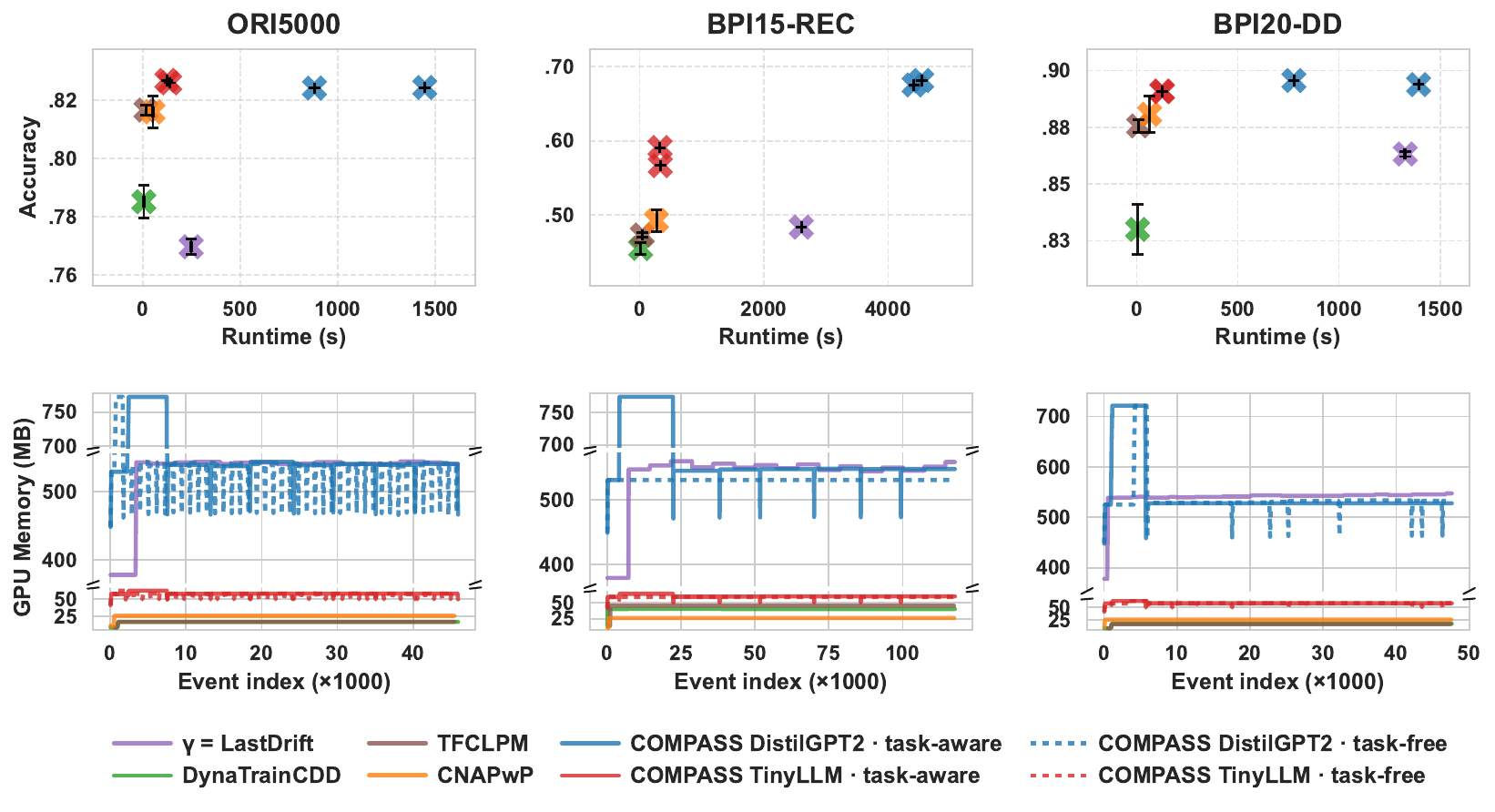}
    \caption{Accuracy versus runtime (top) and memory consumption (bottom) over the event stream  for ORI5000, BPI15-REC and BPI20-DD. COMPASS shows state-of-the-art accuracy with acceptable computational overhead.}
    \label{fig:computational_efficiency}
\end{figure}

Fig.~\ref{fig:computational_efficiency} plots average accuracy against runtime and memory consumption for ORI5000, BPI15-REC and BPI20-DD. COMPASS DistilGPT2 achieves its accuracy gains at a moderate computational cost. It is slower than lightweight baselines such as TFCLPM~\cite{TFCLPM}, DynaTrainCDD~\cite{DynaTrainCDD} and CNAPwP~\cite{CNAPWP} due to the backbone size and SVD-based consolidation steps at task boundaries, but runtime is kept well within practical bounds. Notably, $\gamma$ = LastDrift~\cite{EDBN} incurs comparable or higher runtime on several datasets due to its periodic full retraining, yet consistently underperforms COMPASS in accuracy. Memory usage shows a brief peak at stream onset, caused by the one-time extraction of $W_p$ via SVD. From that point, memory usage remains stable across the stream, as the rank cap on $M_t'$ bounds subspace growth and prevents unbounded memory accumulation regardless of how many tasks are encountered. The Tiny-LLM backbone variant offers a particularly favorable trade-off, approaching DistilGPT2 accuracy at substantially lower runtime and memory footprint, making it an attractive option for resource-constrained deployments. Together, these results show that the accuracy gains of COMPASS come with acceptable computational overhead. Full runtime results can be found in the GitHub repository.

\begin{wraptable}[10]{r}{0.32\textwidth}
\vspace{-4.5em}
\caption{Ablation study using accuracy for our components (TF = Task-Free).}
\label{tab:ablation}
\setlength{\tabcolsep}{4pt}
\renewcommand{\arraystretch}{1}
\centering
\begin{tabular}{cccc r}
\toprule
$W_p$ & $M_t$ & \small\textit{TF} & \small\textbf{Avg.} \\
\midrule
$\times$   & $\times$   & \checkmark & .735\textcolor{gray}{\tiny$\pm$.002} \\
$\times$   & $\times$   & $\times$   & .758\textcolor{gray}{\tiny$\pm$.003} \\
\midrule
\checkmark & $\times$   & \checkmark & .721\textcolor{gray}{\tiny$\pm$.004} \\
\checkmark & $\times$   & $\times$   & .747\textcolor{gray}{\tiny$\pm$.002} \\
\midrule
\checkmark & \checkmark & \checkmark & \textbf{.802\textcolor{gray}{\tiny$\pm$.004}*} \\
\checkmark & \checkmark & $\times$   & \textbf{.802\textcolor{gray}{\tiny$\pm$.003}*} \\
\bottomrule
\end{tabular}
\end{wraptable}

\subsubsection{Ablation Study.}
Table~\ref{tab:ablation} reports the contribution of each component averaged across all nine evaluated datasets. Removing both $W_p$ and $M_t$ reduces average accuracy to .735 (task-free) and .758 (task-aware), establishing the cost of using vanilla LoRA without any subspace protection. Adding $W_p$ alone does not improve over this baseline and slightly hurts performance (.721/.747), suggesting that anchoring the pre-trained subspace without also protecting task-specific directions is insufficient and may unnecessarily constrain plasticity. The full combination of both $W_p$ and $M_t$ recovers the best performance (.802 in both settings), demonstrating that the two components are complementary and jointly necessary. Notably, the task-free and task-aware variants achieve identical average accuracy, confirming that the plateau detection mechanism introduces no meaningful performance penalty compared to oracle boundaries.

\section{Conclusion and Future Work}\label{sec:conclusion}
OCL for PPM has long been constrained by the cold-start problem and the stability-plasticity dilemma, with existing methods rebuilding task-specific networks from scratch at every drift point. COMPASS breaks from this paradigm by continuously fine-tuning a FM via a growing orthogonal knowledge subspace, ensuring that neither important pre-trained sequence understanding nor previously learned process behaviors are overwritten as the stream evolves. Evaluated on nine event streams, COMPASS consistently outperforms five competitors and baselines, with gains of up to 19\% exhibiting recurrent drifts and complex, long-running cases. The task-free variant matches oracle performance across all evaluated datasets, making COMPASS practical for real-world deployments where task boundaries are usually difficult to obtain. The limitations of this work include that the semantic transfer of language pretraining into integer-tokenized activities remains untested, and evaluation is limited to next activity prediction only. Future work includes exploring online or adaptive hyperparameter tuning strategies, extending COMPASS to a multimodal setup and incorporating richer event attributes such as resources.

\bibliographystyle{splncs04}
\bibliography{sn-bibliography}

\end{document}